\documentclass[letterpaper, 10 pt, conference]{ieeeconf}  

\IEEEoverridecommandlockouts                              

\usepackage{amsmath,amsfonts}
\usepackage{algorithm}
\usepackage{mathtools}
\usepackage{array}
\usepackage{textcomp}
\usepackage{stfloats}
\usepackage{url}
\usepackage{verbatim}
\usepackage{graphicx}
\usepackage{subfigure}
\usepackage[font=small]{caption}
\usepackage{algpseudocode}
\usepackage{tablefootnote}
\usepackage{threeparttable}
\usepackage{cite}
\usepackage{hyperref}
\hypersetup{colorlinks=true, linkcolor=blue, anchorcolor=blue, citecolor=blue}
\usepackage[utf8]{inputenc} 
\usepackage[T1]{fontenc}    
\usepackage{url}            
\usepackage{booktabs}       
\usepackage{nicefrac}       
\usepackage{microtype}      
\usepackage{longfbox}
\usepackage{amssymb}
\usepackage{amsmath}   
\usepackage{bbm}
\usepackage{wrapfig}
\usepackage{multirow}
\usepackage{makecell}
\usepackage{parnotes}
\usepackage{cancel}
\usepackage{float}
\usepackage{chngcntr}
\usepackage{apptools}
\usepackage{framed} 
\usepackage[most]{tcolorbox}
\usepackage[algo2e,algoruled,boxed,lined,norelsize]{algorithm2e} 
\usepackage{amsmath} 
\usepackage{amssymb}  

\usepackage{graphicx}
\usepackage{hyperref}

\DeclareMathOperator{\sdf}{\texttt{\textbf{SDF}}}

\newcommand{\ourmethod}{TwinTrack}

\title{\LARGE \bf
TwinTrack: Bridging Vision and Contact Physics \\ for Real-Time Tracking of Unknown  Objects in Contact-Rich Scenes
\vspace{-10pt}
}

\makeatletter
\g@addto@macro\@maketitle{
\setcounter{figure}{0}
  \vspace{-10pt}
  \begin{figure}[H]
  \setlength{\linewidth}{\textwidth}
  \setlength{\hsize}{\textwidth}
  \centering
  \includegraphics[width=\textwidth]{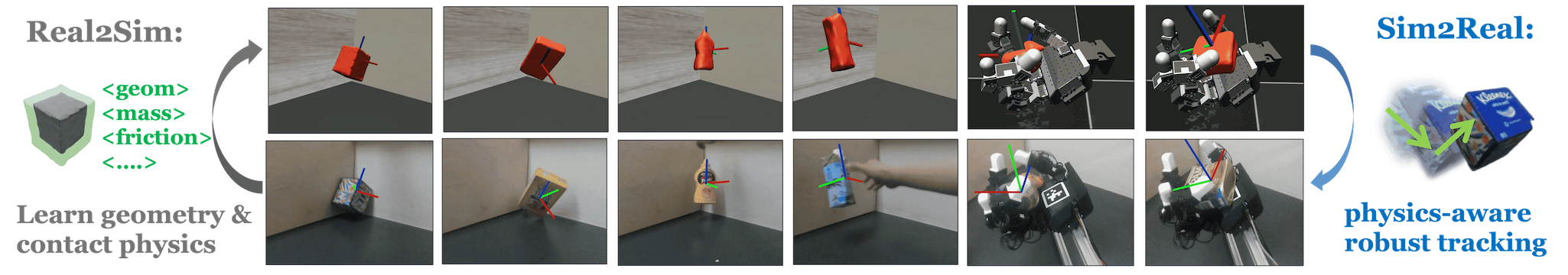}
  \caption{\textbf{\ourmethod} is a physics-aware perception framework
  for robust, real-time tracking of unseen, dynamic objects in
  contact-rich scenes with motion blur
  (Cols. 1–4) and heavy occlusion (Cols. 5–6). The framework integrates two key
  components: \textbf{Real2Sim}, which combines the complementary strengths of
  vision and contact dynamics to jointly estimate an object’s geometry
  and physical properties; and \textbf{Sim2Real}, which performs robust object
  pose estimation through adaptive fusion of visual tracking and contact
  simulation. 
  \href{https://irislab.tech/TwinTrack-webpage/}{video link}
  }
  \label{teaser_fig}
  \end{figure}
  \vspace{-25pt}
}

\author{Wen Yang$^{1,2}$, Zhixian Xie$^{1}$, Yiting Wang$^{1}$, 
Abhijit Tadepalli$^{1}$, Heni Ben Amor$^{3}$, 
Shan Lin$^{4}$, Wanxin Jin$^{1*}$%
\thanks{$^{1}$School for Engineering of Matter, Transport, and Energy.}
\thanks{$^{3}$School of Computing and Augmented Intelligence.}
\thanks{$^{4}$School of Electrical, Computer and Energy Engineering.}
\thanks{All authors are with Arizona State University, Tempe, AZ 85281, USA. Wen Yang$^{2}$ is also with Department of Robotics, Mohamed bin Zayed University of Artificial Intelligence (MBZUAI), Abu Dhabi, UAE.}
\thanks{*Corresponding author: Wanxin Jin (wanxin.jin@asu.edu).}
}

\begin{document}

\maketitle

\begin{abstract}

Real-time tracking of previously unseen, highly dynamic objects in contact-rich scenes, such as during dexterous in-hand manipulation, remains a major challenge. Pure vision-based approaches often fail under heavy occlusions due to frequent contact interactions and motion blur caused by abrupt impacts. We propose \emph{\ourmethod{}}, a physics-aware perception system that enables robust, real-time 6-DoF pose tracking of unknown dynamic objects in contact-rich scenes by leveraging contact physics cues. At its core, \emph{\ourmethod{}} integrates Real2Sim and Sim2Real. Real2Sim combines vision and contact physics to jointly estimate object geometry and physical properties: an initial reconstruction is obtained from vision, then refined by learning a geometry residual and simultaneously estimating physical parameters (e.g., mass, inertia, and friction) based on contact dynamics consistency. Sim2Real achieves robust pose estimation by adaptively fusing a visual tracker with predictions from the updated contact dynamics. \emph{\ourmethod{}} is implemented on a GPU-accelerated, customized MJX engine to guarantee real-time performance. We evaluate our method on two contact-rich scenarios: object falling with environmental contacts and multi-fingered in-hand manipulation. Results show that, compared to baselines, \emph{\ourmethod{}} delivers significantly more robust, accurate, and real-time  tracking in these challenging settings, with tracking speeds above 20 Hz.  \href{https://irislab.tech/TwinTrack-webpage/}{Project page}

\end{abstract}

\section{INTRODUCTION}

Real-time 6-DoF pose tracking of previously unseen, highly dynamic objects in contact-rich scenes is fundamental to robotic manipulation \cite{okamura2000overview,rajeswaran2017learning} and  scene understanding \cite{pfaff2025scalable}. Contact-rich scenes typically involve frequent object-robot-environment interactions and abrupt motion changes, such as  jumps in the object velocity profile, causing large occlusions and motion blur. Pure vision-based tracking \cite{wen2023bundlesdf} struggles with the incomplete and/or degraded observations.

Integrating contact physics cues \cite{pfeiffer2000multibody} can overcome these challenges. Simulated contact dynamics, when aligned with reality, anticipate object poses in highly occluded frames and compensate for degraded visual tracking. Furthermore, incorporating contact dynamics also bridges the gap between visual perception and digital physics simulation, enabling downstream applications such as model-based dexterous manipulation control \cite{contactsdf, taskdrivenmrdm,jin2024complementarity}.

Building physics-informed perception system from monocular RGB-D vision for unknown objects is challenging with real-time constraints. The system must jointly estimate object pose, geometry, and physical properties (e.g., mass, inertia, friction coefficients) from RGB-D inputs. The high-dimensional optimization space requires a modular optimization pipeline with GPU-accelerated implementation. Specifically, three major challenges need to be addressed: (i) Efficiently integrating the complementary benefits of   vision and contact dynamics cues for geometry estimation. Vision provides dense geometric reconstructions, but these are often corrupted by noise and occlusion. In contrast, contact events imply  highly accurate geometric information but are sparse and intermittent \cite{pmlr-v155-pfrommer21a}. (ii) Contact interactions introduce non-smooth motion, which challenges gradient-based  learning and optimization \cite{suh2022differentiable}. (iii) Real-time performance requires highly efficient implementation.

To address these challenges, we propose \emph{\ourmethod{}} (Fig. \ref{teaser_fig}), a physics-aware perception framework that bridges vision and contact dynamics for robust, real-time 6-DoF tracking of  dynamic unknown objects in contact-rich environments. \ourmethod{} takes monocular RGB-D images and robot proprioceptive data as input and outputs real-time 6-DoF object poses that are resilient to occlusions and degraded vision observations. The core contributions of   \ourmethod{} include: 

\emph{(I) An architecture that closes the loop between Real2Sim and Sim2Real.} Real2Sim reconstructs object geometry and aligns a contact dynamics simulator with temporal RGB-D observations. Sim2Real then leverages the aligned simulator to perform physics-aware 6-DoF pose tracking that remains robust under occlusions and degraded visual input.

\emph{(II) Geometry reconstructed from RGB-D vision and refined by contact dynamics.} In Real2Sim, a dense  geometry representation of the object is first extracted from pure vision. Due to noise and occlusion, the geometry is often inaccurate or incomplete. To address this, we propose refining it by learning a  geometry residual in the alignment of contact dynamics simulator, together with estimating  physical properties of the contact dynamics. 
This two-phase process enables effective integration of visual and reliable contact-dynamics cues to derive contact-consistent geometry.

\emph{(III) Real-time implementation and optimization.} Real2Sim (slow backend) and Sim2Real (fast frontend) are implemented separately for real-time tracking performance.  To handle the non-smooth  contact dynamics in Real2Sim optimization, we use GPU-parallelized, sampling-based optimization. 
We developed a deeply customized MJX-based physics engine \cite{mjxgithub}  for collision detection and contact dynamics alignment.

We evaluate \ourmethod{} in two challenging contact-rich scenarios: object free-fall and in-hand manipulation, both involving highly dynamic motion, frequent occlusions, and contact impacts. Compared with the state-of-the-art visual-only tracker \cite{wen2024foundationpose}, \ourmethod{} achieves physically plausible and robust tracking across diverse real-world objects while maintaining real-time performance (>20Hz).

\section{Related Work}\label{sec.related_works}

\subsection{Visual Tracking and Geometry Reconstruction}
Vision-based estimation of object pose and geometry  has been a long-standing focus in computer vision. Methods with strong priors (CAD models, template matching) \cite{labbe2020cosypose, labbe2022megapose, wang2019normalized} achieve high accuracy but lack generalization to unseen objects. More recent approaches aim to track previously unseen objects without any prior knowledge, either by integrating pre-trained image descriptors~\cite{wen2021bundletrack} or by jointly learning 3D representations during tracking via neural fields~\cite{wen2023bundlesdf, wen2024foundationpose}.

While vision-based approaches perform well in occlusion-free and moderately dynamic conditions \cite{wen2021bundletrack, wen2023bundlesdf, wen2024foundationpose}, they struggle in highly dynamic, contact-rich scenarios involving fast motion, abrupt changes, and frequent occlusion. Visual observations become degraded or incomplete, leading to unreliable tracking. Object geometry from visual data may be uncertain due to noise, causing inconsistencies in motion prediction \cite{pmlr-v155-pfrommer21a}. This underscores the need to incorporate physics knowledge (contact dynamics) to enhance robustness and ensure physically plausible estimates. 

\subsection{Physics-Informed Tracking and Reconstruction}
Physics-informed tracking incorporates physics priors to enforce physical plausibility. Human motion capture (e.g., \cite{gartner2022differentiable}) uses simulation-based constraints and soft physics priors for improved realism, but often assumes known physical properties limiting generalization. More recent approaches (e.g., \cite{zhang2023slomo}) focus on jointly learning geometry and physical parameters by minimizing simulation/prediction loss from differentiable simulators. For example, \cite{pfaff2025scalable} proposes using Real2Sim to extract collision geometry from robot sensors and cameras, while \cite{jiang2025phystwin} reconstructs interactive digital twins from object interaction videos. However, their scenarios involve relatively slow and less occluded motion. In contact-rich scenes with high occlusion and motion blur, visual geometry may be inaccurate, degrading dynamics learning. Thus, \cite{bianchini2025vysics} proposes combining visual and contact cues for geometry inference under occlusion, but they focus on reconstruction rather than real-time tracking.

\begin{figure*}[ht]
  \centering
  \includegraphics[scale=0.38]{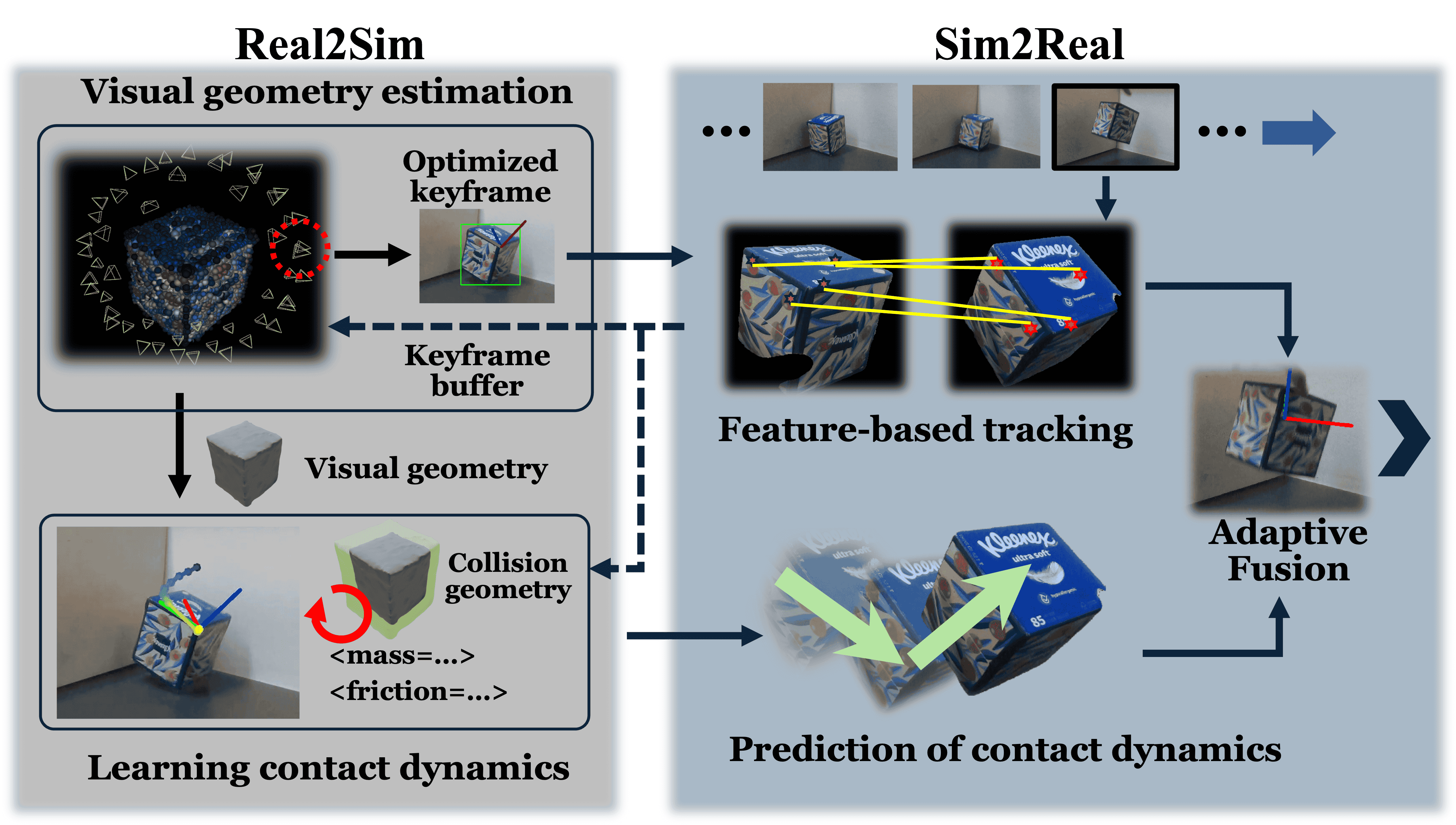}
  \caption{Overview of \textbf{\ourmethod{}} framework. Our framework includes two main components: \textbf{Real2Sim} for learning object geometry and simulation-ready contact physics, and \textbf{Sim2Real} for physics-aware real-time  pose tracking. In \textbf{Real2Sim},  object visual geometry, represented as the Gaussian Splats, is obtained from a selection of RGB-D keyframes (Sec.~\ref{sec:geom_reconstruction}). The obtained geometry is  updated in the next stage of learning contact dynamics, together with identifying other physical properties (Sec.~\ref{sec:dynamics_adaptation}). In \textbf{Sim2Real}, feature-based tracking is performed for each new frame (Sec.~\ref{sec:frontend_visual_correspondence}) relative to  optimized keyframes from \textbf{Real2Sim}; meanwhile the  contact physics simulation also predicts the current object pose. The final object pose is an adaptive fusion of both visual tracker and  simulation (Sec.~\ref{sec:contact_guided_frontend_tracking}).
  }
  \vspace{-15pt}
  \label{fig:pipeline}
\end{figure*}

\subsection{Learning Contact Dynamics}

Accurate physics models are essential for physics-informed perception \cite{jiang2025phystwin}. However, real-world contact interactions are inherently non-smooth with stiff transitions, posing learning challenges. Standard neural networks struggle with discontinuities due to smooth bias \cite{suh2022differentiable}. Differentiable simulators (e.g., MJX \cite{mjxgithub}) provide gradients that are often discontinuous and local, limiting global optimization \cite{suh2022differentiable}. Studies \cite{suh2022differentiable, contactsdf,jin2024complementarity} relax stiff constraints for smooth gradients. Our method avoids contact gradients by adopting sampling-based optimization. Furthermore, built on the neural { signed distance function (SDF)} representation of contact geometry in the simulator, we propose learning geometry residuals to refine the functional geometry used in contact dynamics, enhancing simulation-reality alignment.

\section{Problem Statement and  \ourmethod{} Overview}\label{sec:Approach}

We consider RGB-D image sequences of a contact-rich dynamic scene where an unknown object frequently interacts with robots or/and environments. Denote RGB and depth image at frame $k$ as $(I_k,D_k)$. \ourmethod{} takes RGB-D images and robot proprioceptive data (when available) as input. The goal is to estimate the 6-DoF object pose $\mathbf{q}_k$ in real-time despite degraded visual observations from occlusion and motion blur due to contact impacts.

{As shown in Fig.~\ref{fig:pipeline}, \ourmethod{} handles the problem by a backend--frontend loop between Real2Sim and Sim2Real. Sim2Real runs online and produces a per-frame pose estimate for the current observations. Real2Sim runs asynchronously on accumulated observational history to update the object model and keyframes that Sim2Real can reliably use. The two components exchange information continuously: Sim2Real provides keyframes and tracking history, while Real2Sim returns updated keyframe anchors and an improved contact dynamics model.
At the beginning, the system relies mainly on vision; as Real2Sim becomes reliable, the learned physical model is progressively incorporated to stabilize tracking under occlusion. The details will be presented in Section \ref{sec.real2sim} and \ref{sec.Sim2Real}.}

\section{Real2Sim}\label{sec.real2sim}

This section describes the Real2Sim (backend), whose goal is to produce (i) optimized keyframes with consistent poses and (ii) a simulation-ready contact dynamics model.
Real2Sim is a two-phase process: the object visual geometry is first estimated using RGB-D keyframes from a keyframe buffer (Sec.~\ref{sec:geom_reconstruction});  we then fine the geometry  to align better with observed contact behavior and  identify the  physical parameters (Sec.~\ref{sec:dynamics_adaptation}) using tracking data provided by Sim2Real (Sec.~\ref{sec.Sim2Real}). Inputs to Real2Sim: keyframe buffer and historical tracking data from Sim2Real. Outputs: the simulation-ready contact dynamics and pose-optimized keyframes.

\subsection{Visual Geometry Reconstruction}
\vspace{-5pt}
\label{sec:geom_reconstruction}
\begin{figure}[ht]
  \centering
  \includegraphics[scale=0.35]{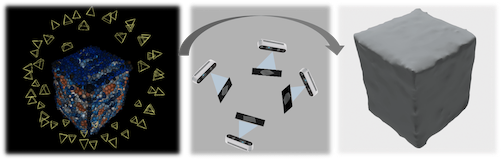}
  \caption{Visual geometry reconstruction. Left: joint optimization of Gaussian Splatting (GS) model $\mathcal{G}_{\text{gs}}$ and keyframe poses. Middle: depth rendering from the  obtained GS model. Right: learning neural SDF $\sdf_{{\text{vision}}}$ from the rendered depth images.}
  \vspace{-5pt}
  \label{fig:pose_geometry_learning_diagram}
\end{figure}

This goal is to jointly reconstruct the visual geometry of the unknown object from keyframes and simultaneously optimize the keyframe poses. The reconstructed geometry will be refined in contact dynamics learning (Sec.~\ref{sec:dynamics_adaptation}), while the optimized keyframes serve as pose anchors for visual tracking in Sim2Real (Sec. \ref{sec:frontend_visual_correspondence}).

As  in Fig. \ref{fig:pose_geometry_learning_diagram},
we use the Gaussian Splatting (GS) primitives  $\mathcal{G}_{\text{gs}}$ \cite{kerbl2023_3dgs} to represent the spatial model of the object, jointly encoding geometry and appearance. Following \cite{matsuki2024gaussian}, we jointly optimize the GS model $\mathcal{G}_{\text{gs}}$ and  the poses of keyframes $\{\mathbf{q}_i\}_{i\in\mathcal{K}}$ from  a keyframe buffer $\mathcal{K}$ (detailed in Section \ref{sec:frontend_visual_correspondence}) based on the following loss
\begin{equation}
\begin{aligned}
    \min_{\{\mathbf{q}_i\}_{i \in \mathcal{K}},\, \mathcal{G}_{\text{gs}}} 
    \sum\nolimits_{i \in \mathcal{K}} \lambda_\text{pho} E_i^{\text{pho}} + (1-\lambda_\text{pho})E_i^{\text{geo}}
\end{aligned}
\end{equation}
with $E_i^{\text{pho}}$ the photometric residual and $E_i^{\text{geo}}$ the geometric loss for the keyframe $i$, respectively defined as
\begin{equation}
    \begin{aligned}
E_i^{\text{pho}}&=||I(\mathcal{G}_{\text{gs}},\mathbf{T}(\mathbf{q}_i)) - I_i||_1,\\
E_i^{\text{geo}}&=||D(\mathcal{G}_{\text{gs}},\mathbf{T}(\mathbf{q}_i)) - D_i||_1,
    \end{aligned}
\end{equation}
where $I(\mathcal{G}_{\text{gs}},\mathbf{T}(\mathbf{q}_i))$ renders the Gaussians $\mathcal{G}_{\text{gs}}$ for pose $\mathbf{q}_i$ by $\alpha$-blending; $D(\mathcal{G}_{\text{gs}},\mathbf{T}(\mathbf{q}_i))$ is the  depth rendering; 
$I_i$ and $D_i$ are the observed image and depth of keyframe $i$; and $\mathbf{T}(\cdot)$ converts pose to transformation matrix.
For the optimization  and  implementation details, refer to   \cite{matsuki2024gaussian}.

The obtained  $\mathcal{G}_{\text{gs}}$ can be difficult to be directly integrated in physics engine. Thus, as the final step in this phase, we learn {a neural signed distance function (SDF) representation}, denoted as 
\begin{equation}\label{equ.vision_sdf}
    \phi=\sdf_{{\text{vision}}}(\mathbf{p}), \quad \mathbf{p}\in\mathbb{R}^3
\end{equation}
\noindent via the ray-casting sampling on the rendered depth from the  GS model $\mathcal{G}_{\text{gs}}$, as  in Fig. \ref{fig:pose_geometry_learning_diagram}. The obtained $\sdf_{{\text{vision}}}$ will then be  used in the  contact dynamics (simulator) in Section \ref{sec:dynamics_adaptation}.
 
\subsection{Contact Dynamics and Geometry Compensating}
\label{sec:dynamics_adaptation}
Our contact dynamics model is built on the MJX physics engine \cite{todorov2012mujoco, mjxgithub, brax2021github} with substantially redesigned collision detection routine  to integrate the obtained neural SDF from the previous phase. The JAX implementation of the MJX physics engine  enables GPU-parallel simulation and sampling-based optimization in this subsection.
In generic notation, the contact dynamics~\cite{mjxgithub} can be modeled as
\begin{equation}\label{equ:mjx}
\begin{aligned}
&\mathbf{M}(\mathbf{q}_{k}) (\mathbf{v}_{k+1}-\mathbf{v}_{k}) 
= h\mathbf{c}(\mathbf{q}_{k}, \mathbf{v}_{k}) 
+ h\mathbf{u}_{k} 
+ \sum\nolimits_{i}
\mathbf{J}_i(\mathbf{q}_{k})^\top \boldsymbol{\lambda}_i,
\\
&\boldsymbol{0} \leq \boldsymbol{\lambda}_i \perp 
\boldsymbol{\phi}_i(\mathbf{q}_{k+1}) \geq \boldsymbol{0}, 
\quad i \in \mathcal{C}(\mathbf{q}_{k}, \sdf_{\text{collision}}).
\end{aligned}
\end{equation}
\noindent Here $h$ is the time interval, $\mathbf{q}_{k} \in \mathbb{R}^{n_q}$ and 
$\mathbf{v}_{k} \in \mathbb{R}^{n_v}$ are the generalized configuration and velocity of all degrees of freedom (object and robot),
the inertia matrix $\mathbf{M}(\mathbf{q}_k) \in \mathbb{R}^{n_v \times n_v}$, 
the bias force  $\mathbf{c}(\mathbf{q}_{k}, \mathbf{v}_{k}) \in \mathbb{R}^{n_v}$ includes gravity and inertial effects, applied force is $\mathbf{u}_{k} \in \mathbb{R}^{n_v}$. 
$\boldsymbol{\lambda}_i \in \mathbb{R}^{m_i}$ is the $i$-th frictional contact impulse with corresponding contact Jacobian $\mathbf{J}_i(\mathbf{q}_{k}) \in \mathbb{R}^{m_i \times n_v}$.
$\boldsymbol{\phi}_i(\mathbf{q}_{k+1}) \in \mathbb{R}^{m_i}$ defines the general contact distance (i.e., gaps in normal direction and for dual cone faces of friction cone), with  $m_i$ being the contact dimension.  
$\mathcal{C}(\mathbf{q}_{k}, \sdf_{\text{collision}})$ is the collision detection module which returns the active contact set based on the collision SDF  $\sdf_{\text{collision}}$ queries. 
The second inequality denotes complementarity constraint enforcing $\boldsymbol{\lambda}_i \ge \boldsymbol{0}$, $\boldsymbol{\phi}_i(\mathbf{q}_{k+1}) \ge \boldsymbol{0}$, and $\boldsymbol{\lambda}_i^\top \boldsymbol{\phi}_i(\mathbf{q}_{k+1}) = \boldsymbol{0}$. 
The configuration update follows the semi-implicit Euler step $\mathbf{q}_{k+1} = \mathbf{q}_k \oplus h\mathbf{v}_{k+1}$, with $\oplus$ including quaternion integration.

\begin{figure}[ht]
  \centering
  \includegraphics[width=0.5\textwidth]{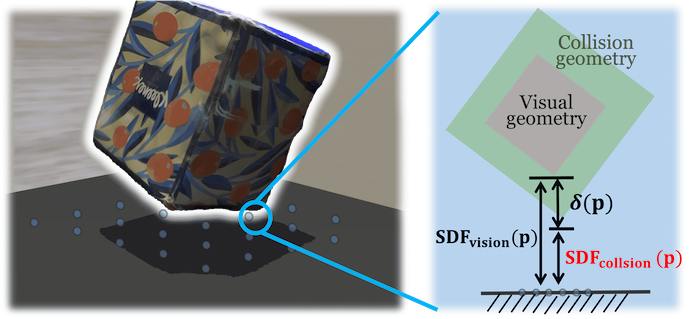}
  \caption{Collision geometry is a sum of visual geometry representation and a learnable  geometry residual.}
  \vspace{-5pt}
  \label{fig:diff_sdf_margin_schematic}
\end{figure}

In our method, we modify the following aspects of the MJX physics engine to better integrate with neural geometry and contact dynamics learning. 

\paragraph{Geometry Residual} In each physics simulation step (\ref{equ:mjx}), collision detection routine  requires querying \emph{collision geometry} $\sdf_{{\text{collision}}}$ to obtain necessary collision information. Visual geometry $\sdf_{\text{vision}}$ reconstructed in (\ref{equ.vision_sdf}) can be imperfect to be directly   used as collision SDF $\sdf_{{\text{collision}}}$, due to noisy and/or partial vision observation. Thus, we propose to obtaining the collision SDF  $\sdf_{{\text{collision}}}$ by refining the visual geometry with a residual term:
\begin{equation}\label{equ:compensation}
     \sdf_{{\text{collision}}}(\mathbf{p})=\sdf_{{\text{vision}}}(\mathbf{p})-\delta(\mathbf{p}).
\end{equation}
The  term $\delta(\mathbf{p})$ is called \emph{geometry residual}, which  compensates for visual geometry inaccuracies. The geometry residual $\delta(\mathbf{p})$ is learned along with other physics parameters (mass, inertia, friction coefficient) in (\ref{equ:mjx}) in Sec. \ref{sec:learn_dynamics}.

\paragraph{Collision Detection with Neural SDFs}
MJX physics engine mainly supports convex meshes and primitive shapes for collision detection. To handle arbitrary non-convex neural geometry represented by $\sdf_{{\text{collision}}}$, we customize MJX's collision detection routine following \cite{contactsdf}. Specifically, collision detection with SDF geometry involves querying $\sdf_{\text{collision}}(\mathbf{p})$ with surface points $\mathbf{p}$ sampled from surrounding entities (ground, walls, robot fingers) as shown in Fig. \ref{fig:diff_sdf_margin_schematic}. JAX-GPU acceleration enables parallel distance queries with reasonably low computational overhead.

\subsection{Learning Contact Dynamics}\label{sec:learn_dynamics}
Learning contact dynamics  involves estimating  physical parameters in (\ref{equ:mjx}), 
\begin{equation}\label{equ:theta}
    \boldsymbol\theta=\left\{
    m_{
    \text{obj}}, 
    \text{diag}[I_x,I_y,I_z]_{\text{obj}}, 
    \mu, \delta
    \right\},
\end{equation}
including object mass, diagonalized inertia matrix, friction coefficient, and collision geometry residual (\ref{equ:compensation}). We update all those parameters  by minimizing the one-step prediction (simulation) loss of the contact dynamics:
\begin{equation}\label{equ:mjx_update}
\mathcal{L}(\boldsymbol{\theta})=
\sum_{(\mathbf{q}_k, \mathbf{v}_{k}, 
\mathbf{q}_{k{+}1}) \in \mathcal{D}_{\text{tracking}}} \left\| \mathrm{sim}_{\boldsymbol{\theta}}(\mathbf{q}_{k}, \mathbf{v}_{k}
) - \mathbf{q}_{k+1} \right\|_{W}^2,
\end{equation}
where $\mathcal{D}_{\text{tracking}}$ is history tracking data from Sim2Real (Section \ref{sec.Sim2Real});  $\mathrm{sim}$ denotes one-step simulation based on the contact dynamics model (\ref{equ:mjx}) with  integration $\mathbf{q}_{k+1} {=} \mathbf{q}_k \oplus h\mathbf{v}_{k+1}$; and the weight matrix $W$ balances translation and rotation losses.

Note that for geometry residual learning in (\ref{equ:mjx_update}), we adopt a constant residual  $\delta$ rather than a more expressive residual function $\delta(\mathbf{p})$. This is because the contact events are inherently sparse and intermittent,  fine-tuning an expressive residual function (e.g., neural networks) can be unstable and prone to overfitting, although we acknowledge that this could be addressed by using regularization and careful parameterization. We will explore them in our future work. Here, a constant residual $\delta$  means correcting the visual geometry  by uniformly expanding or shrinking  by a margin. This provides a lightweight yet practically effective mechanism (in our cases) to refine contact behavior. 

Due to the non-smoothness of contact dynamics (\ref{equ:mjx}), the  gradient-based optimization is hard to apply to (\ref{equ:mjx_update}). While the MJX engine supports differentiable simulation but the gradients are limited in local contact modes, not useful to provide global optimization guidance \cite{suh2022differentiable}. On the other hand, MJX enables highly parallelized simulations across randomized environments, making it suitable for sample-based optimization. Thus, we use a variant of  cross-entropy method (CEM) \cite{rubinstein2004cross} to solve (\ref{equ:mjx_update}), where  unlike standard CEM methods that update based on elite samples, we use softmax-weighted average over all samples (similar to \cite{mppi_icra}), Specifically, let $j$ denote learning iteration, and  $N$ the sample size at each iteration.  The particle $\boldsymbol\theta^{(j+1)}_i$, $i = 1, 2, …, N$, at the $(j+1)$-th iteration is sampled from $\boldsymbol\theta^{(j+1)}_i \sim \mathcal{N}(\boldsymbol\mu^{(j+1)}, \boldsymbol\Sigma)$, with
\begin{equation}
    \boldsymbol\mu^{(j+1)}=\frac{\sum_{i=1}^N \exp{\left(\gamma \mathcal{L}(\boldsymbol{\theta}_i^{(j)})\right)}\boldsymbol{\theta}_i^{(j)}}{\sum_{i=1}^N \exp{\left(\gamma \mathcal{L}(\boldsymbol{\theta}_i^{(j)})\right)}}.
\end{equation}
where $\gamma$ is the temperature (hyperparameter), and $\boldsymbol\Sigma$ is set as a diagonal  constant matrix with entries depending on physical unit scales. This update is similar to the MPPI method \cite{williams2017information}.

\section{Sim2Real}\label{sec.Sim2Real}
\vspace{-5pt}
This section presents Sim2Real (frontend), an online module that estimates  object pose at each frame by combining two complementary cues:
a visual pose estimate obtained by matching the current frame to nearby optimized keyframes, and a simulation state from the learned contact dynamics.
We introduce a feature-based visual tracker in Sec.~\ref{sec:frontend_visual_correspondence}, and then describe the pose optimization and adaptive fusion in Sec.~\ref{sec:contact_guided_frontend_tracking}.
Besides outputting the real-time pose, Sim2Real maintains a keyframe buffer and logs tracking history, which are used by Real2Sim for continuous updates.

\subsection{Feature-Based Visual Tracker}
\label{sec:frontend_visual_correspondence}
\vspace{-3pt}
Feature-based visual tracker computes relative pose between the current frame and an optimized keyframe (pose anchor) by detecting and matching visual features. The pose anchor is the nearest optimized keyframe from Real2Sim (Sec. ~\ref{sec:geom_reconstruction}). The process follows  \cite{wen2023bundlesdf},  summarized below.

\paragraph{Feature Matching} We segment object region using SAM2 \cite{ravi2024sam2} with prompt given only at the first frame. SuperPoint \cite{superpoint} extracts 3D object features  from the  current frame and reference keyframe. LightGlue \cite{lindenberger2023lightglue} performs feature matching, followed by RANSAC-based pose estimator to filter wrong correspondences. 

\paragraph{Keyframe Buffer}\label{sec:keyframe_selection}
We use keyframes as pose anchors instead of the previous frames to avoid long-term drift. The keyframe is selected following the setup of \cite{wen2021bundletrack, wen2023bundlesdf}, and a  keyframe buffer $\mathcal{K}$ is built to contain multi-view but sparse keyframes, whose poses are continuously optimized in Real2Sim (Sec. \ref{sec:geom_reconstruction}). The first frame is always a keyframe. New frames are added if rotational geodesic distances to existing keyframes exceed threshold \cite{wen2021bundletrack,wen2023bundlesdf}.

\paragraph{Feature-Based Pose Estimation} 
For each new frame, the closest keyframe is selected from the keyframe buffer. Object pose is estimated from matched (and filtered) features between the current frame and selected keyframe. We use a single frame pair instead of multi-keyframe pose graph optimization \cite{wen2021bundletrack,wen2023bundlesdf} for real-time performance, with accuracy refined by fusion with contact physics prediction.

\subsection{Pose Optimization by Adaptive Fusion}
\label{sec:contact_guided_frontend_tracking}
\vspace{-3pt}
While object pose estimation from feature-based tracking and contact dynamics  can be formulated as a Kalman filter \cite{julier1997ekf,wan2000unscented}, we propose a simple yet effective strategy that avoids the complexity of optimization and update. At frame $k$, we estimate the object pose $\mathbf{q}_{k}$  by jointly minimizing visual feature correspondence loss and one-step  prediction loss of the contact dynamics (simulation):
\begin{equation}
\begin{aligned}
    \min_{\mathbf{q}_{k}}\quad 
    w_{k} \mathcal{L}_{\text{vision}}(\mathbf{q}_{k}) + 
    (1-w_{k}) \mathcal{L}_{\text{dynamics}} (\mathbf{q}_{k})
    , 
\end{aligned}
\end{equation}
where  $\mathcal{L}_{\text{vision}} (\mathbf{q}_{k})$ is the  visual feature correspondence loss;  $\mathcal{L}_{\text{dynamics}} (\mathbf{q}_{k})$ is the contact dynamics prediction loss; the weight  $w_k \in [0, 1]$ balances the contributions of the two and will be \emph{adaptively set}  based on the quality of  feature correspondences. We  explain the three terms below.

Visual feature correspondence loss $\mathcal{L}_{\text{vision}}(\mathbf{q}_{k})$ quantifies sum of squared distances of  matched 3D feature points between the current frame and selected keyframes:
\begin{equation}
    \mathcal{L}_{\text{vision}}(\mathbf{q}_{k}) = \sum\nolimits_{j} \left\| \mathbf{T}(\mathbf{q}_{k}) \, \mathbf{p}_{k}^j - \mathbf{T}(\mathbf{q}_{n}) \, \mathbf{p}_n^j \right\|_2^2,
\end{equation}
where $(\mathbf{p}_{k}^j, \mathbf{p}_{n}^j)$ is the $j$-th matched feature pair from current frame pose $\mathbf{q}_k$ and selected keyframe pose $\mathbf{q}_{n}$.

One-step contact dynamics prediction loss is defined as
\begin{equation}
\begin{aligned}
\mathcal{L}_{\text{dynamics}} (\mathbf{q}_{k})=
    \left\|\mathbf{q}_{k} - 
    \mathrm{sim}(\mathbf{q}_{k-1}, \mathbf{v}_{k-1})\right\|_2^2, 
\end{aligned}
\end{equation}
where $\mathrm{sim}$ is (\ref{equ:mjx}) learned in Real2Sim (Section \ref{sec:learn_dynamics}), $\mathbf{q}_{k-1}$ is previous frame's robust pose estimate, and $\mathbf{v}_{k-1}$ is velocity estimated via finite difference on visual pose (if reliable) or from simulator dynamics evolution.

\begin{figure}[ht]
  \centering
  \includegraphics[scale=0.35]{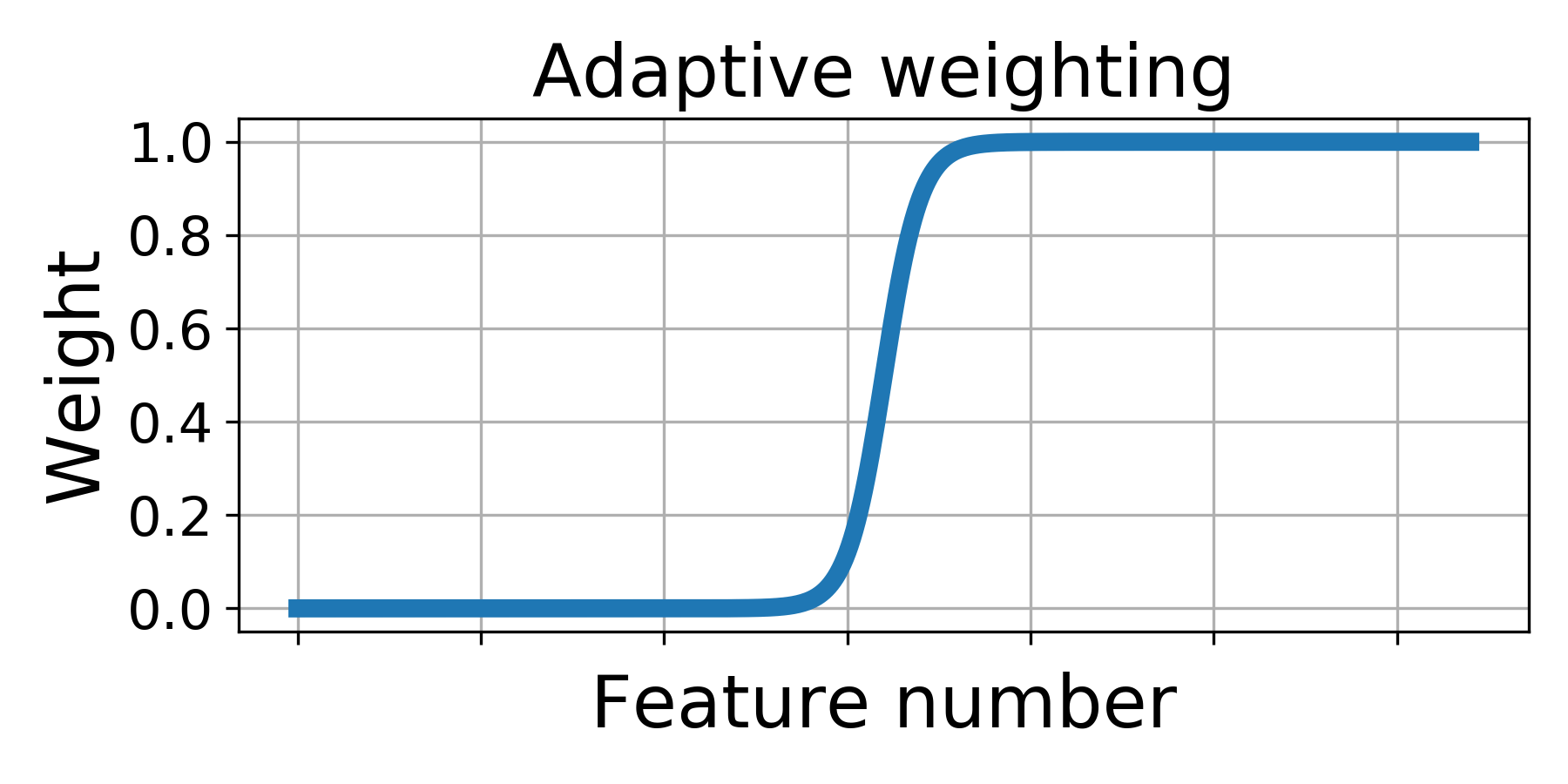}
  \vspace{-5pt}
  \caption{\small Adaptive weighting parameter}
  \vspace{-10pt}
  \label{fig:adaptive_weighting_plot}
\end{figure}

Weight $w_k\in[0,1]$ controls balance between physics-based prediction and feature-based estimation. We set $w_k$ adaptively based on feature matching quality, quantified by number of matched feature pairs after filtering (Section \ref{sec:frontend_visual_correspondence}). Higher reliable feature matches indicate favorable visual conditions (less occlusion and motion blur), so visual estimate gets more weight. We define $w_k$ as sigmoid-like function of matched feature pairs (Fig. \ref{fig:adaptive_weighting_plot}).

\section{Experiments}

We evaluate  \ourmethod{}  across diverse contact-rich scenarios, focusing on its two key components: (1) Real2Sim: how effectively it learns contact dynamics to align with and predict real-world contact-rich motion, and (2) Sim2Real: how much it improves tracking robustness in highly dynamic, contact-rich environments. These evaluations highlight \ourmethod{}’s ability to bridge  sim-to-real gap and enable robust perception.

\vspace{-5pt}
\subsection{Tasks and Dataset}
\vspace{-5pt}
\begin{figure}[ht]
  \centering
  \includegraphics[scale=0.25]{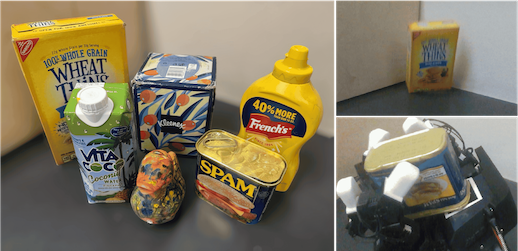}
  \caption{Two contact-rich scenarios with different objects.}
  \label{fig:experiments_scenes}
  \vspace{-10pt}
\end{figure}

Our evaluation centers on two  contact-rich scenarios across various objects (Fig.~\ref{fig:experiments_scenes}):

\textbf{Scenario I: Object falling with rich impacts:} The scene consists of two walls and ground forming a corner with three orthogonal planar surfaces. Objects are released from human hand with initial velocity, undergo free fall, and collide with walls and ground. The scene is impact-rich, and the instantaneous impulses cause abrupt velocity changes of object motion, leading to frequent motion blur. Objects used include snack box, tissue box, milk bottle, Spam can, 3D-printed duck and mustard bottle.

\textbf{Scenario II: In-hand manipulation with occlusion:} LEAP four-fingered robotic hand \cite{shaw2023leaphand} manipulates objects with random policies. The motion involves rich contact between the object and  hand fingers and palm. 
The primary challenge arises from  partial occlusions of the object. The object also undergoes abrupt impacts from  randomly moving fingers. We use the Spam can and 3D-printed duck, as their sizes make them well-suited for handheld interaction.

We collect RGB-D videos using an Intel RealSense D435i camera for both scenarios. In Scenario II, we additionally record the joint states of the robotic hand. For each scenario-object pair, we capture four trajectories under varying motion conditions: three with mild contact interactions (such as planar sliding and lateral shaking) to facilitate the learning of contact dynamics, and one with highly dynamic interactions involving abrupt and rich impacts for robustness evaluation. In Scenario I, data collection begins from the free-motion phase following object release. The object’s initial velocity is estimated with a visual tracker applied to the first few frames.

For ground-truth tracking, we offline reconstruct high-quality textured mesh models using Scaniverse app \cite{scaniverse2021}, then use FoundationPose \cite{wen2024foundationpose} to generate pseudo ground-truth object poses for all video frames.

\begin{figure*}[ht]
  \centering
  \includegraphics[scale=0.10]{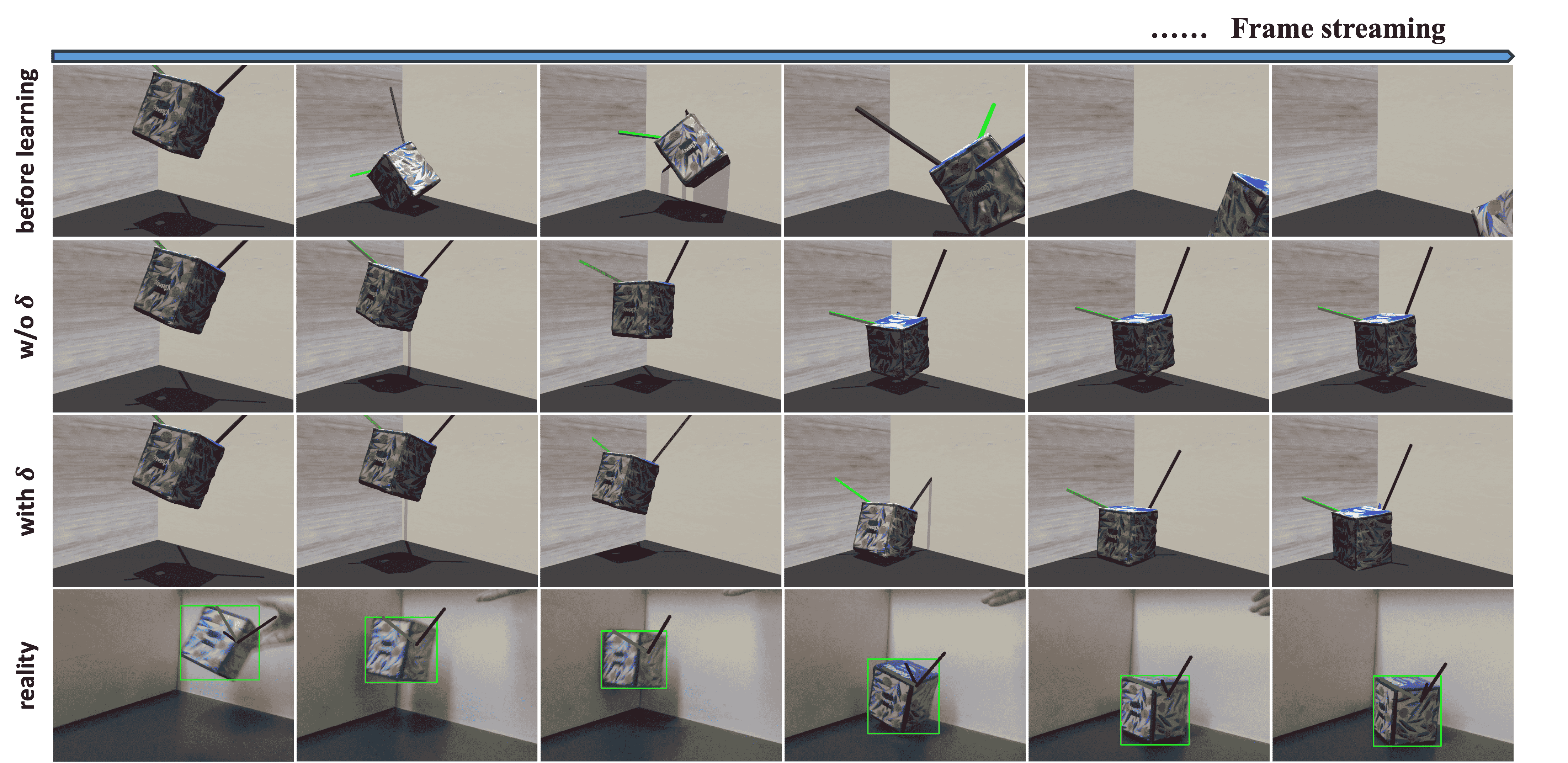}
  \caption{First row: Open-loop simulation using  random physical parameters (prior to Real2Sim).
Second row: Open-loop simulation of the  contact dynamics learned but without geometry refinement, i.e., directly using  visual geometry for simulation.
Third row: Open-loop simulation of the learned contact dynamics, including   geometry refinement. Fourth row: object's actual trajectory in real world.}
  \label{fig:dyn_adapt_tissue_box}
  \vspace{-10pt}
\end{figure*}

\vspace{-1pt}
\subsection{Evaluation of Real2Sim}
\vspace{-5pt}
\begin{figure*}[ht]
  \centering
  \includegraphics[scale=0.57]{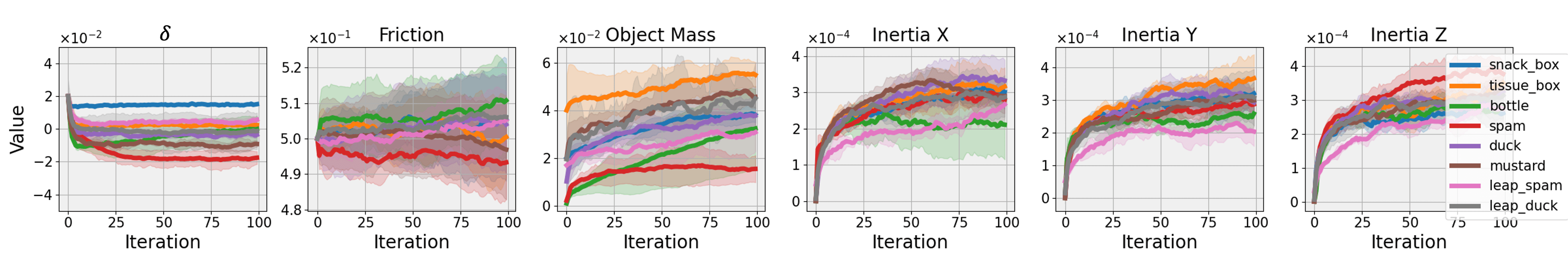}
  \caption{Convergence of all physics parameters in contact dynamics learning for all scenario-object pairs. Each learning runs for 5 trials with different random seeds.  The solid lines represent the mean values, and the shaded areas indicate the standard deviation across the five trials.}
  \vspace{-15pt}
  \label{fig:parameters_learning_one_row}
\end{figure*}

For each scenario-object pair, Real2Sim learns contact dynamics using multiple trajectories with diverse contact interactions: bouncing against walls, sliding along ground, and rich contact in robot hand. 

\begin{table}[htbp]
\centering
\scriptsize
\setlength{\tabcolsep}{4pt}
\renewcommand{\arraystretch}{1.3}
\caption{One-step prediction error (\ref{equ:mjx_update}) ($100\times$ position error + quaternion error) of  contact dynamics model before/after learning.}
\begin{tabular}{c|c c c c c c c c}
\Xhline{1pt}
 & \makecell{\textbf{snack}\\\textbf{box}} & \makecell{\textbf{tissue}\\\textbf{box}} & \makecell{\textbf{bottle}} & \makecell{\textbf{spam}\\} & \textbf{duck} & \textbf{mustard} & \makecell{\textbf{LEAP}\\\textbf{spam}} & 
 \makecell{\textbf{LEAP}\\\textbf{duck}}\\
\hline
Before & 18.78 & 9.57 & 9.82 & 7.24 & 5.13 & 10.18 & 4.18 & 23.03 \\
After w/o $\delta$  & 0.29 & 0.47 & 0.87 & 1.16 & 0.38 & 0.56 & 0.18 & 0.38 \\
After with $\delta$  & 0.20 & 0.22 & 0.70 & 0.43 & 0.21 & 0.15 & 0.13 & 0.28 \\
\Xhline{1pt}
\end{tabular}
\vspace{-15pt}
\label{tab:multi_object_adaptation_loss}
\end{table}

Table~\ref{tab:multi_object_adaptation_loss} shows the prediction error  (\ref{equ:mjx_update}) of the contact dynamics model  before and after learning. "Before": initial contact dynamics model (\ref{equ:mjx}) with random physical parameters  shows substantial gap from reality, with simulated objects exhibiting exaggerated rebounds/unrealistic sliding due to inaccurate mass/inertia parameters. "After w/o $\delta$": learning all physical parameters (\ref{equ:theta}) except  geometry residual $\delta$ (i.e., only using visual geometry) reduces gap but suffers from inaccurate contact behaviors. "After with $\delta$": learning all parameters including $\delta$ further lowers prediction error, indicating better alignment between the contact dynamics model and reality. Learned $\delta$ may be imprecise due to camera calibration errors but remains physically plausible for downstream tracking.

\begin{table}[htbp]
\centering
\scriptsize
\setlength{\tabcolsep}{4pt}
\renewcommand{\arraystretch}{1.3}
\caption{The learned mass [g] of each object versus ground truth.}
\begin{tabular}{c|c c c c c c c c}
\Xhline{1pt} 
 & \makecell{\textbf{snack}\\\textbf{box}} & \makecell{\textbf{tissue}\\\textbf{box}} & \textbf{bottle} & \textbf{spam} & \textbf{duck} & \textbf{mustard} & \makecell{\textbf{LEAP}\\\textbf{spam}} & 
 \makecell{\textbf{LEAP}\\\textbf{duck}}\\
\hline
Mass & 40 & 50 & 24 & 28 & 44 & 65 & 28 & 44 \\
Learned    & 38$\pm$3 & 55$\pm$5 & 32$\pm$10 & 15$\pm$5 & 38$\pm$7 & 46$\pm$5 & 31$\pm$5 & 47$\pm$10 \\
\Xhline{1pt} 
\end{tabular}
\vspace{-10pt}
\label{tab:object_weight_result}
\end{table}

Fig.~\ref{fig:parameters_learning_one_row} shows the learning convergence of all physical parameters for each scenario-object pair (5 trials with different random seeds). Table~\ref{tab:object_weight_result} reports learned mass versus ground-truth. Despite convergence trends, noticeable gaps remain between learned and ground-truth parameters due to: (1) weak observability of parameters like friction/inertia when motion lacks sufficient richness/excitation in corresponding degrees of freedom; (2) MJX's simplified contact handling (point/rigid contact) oversimplifying real-world contact interactions, creating inherent modeling bias: the learning converges to parameters fitting observed motion well despite deviating from true ones. We also note that the learned parameters for Spam/Duck differ between falling and manipulation scenarios. This is likely due to LEAP hand model-hardware mismatch (the hand control parameters not updated in simulation). Additionally, friction and mass do not converge to single consistent values across runs. This may be due to (i) trajectory-dependent excitation of contact and motions, yielding different parameters, and (ii) coupling in optimization, where friction and mass can partially trade off to absorb modeling residuals.

Fig.~\ref{fig:dyn_adapt_tissue_box} shows multi-step open-loop simulation of the contact dynamics model for tissue box falling alongside real-world video (last row "reality"). Initial position/velocity are aligned. First row ("before learning"): randomly initialized contact dynamics model shows highly unrealistic motion (exaggerated rebounds, missed contact). Second row ("w/o $\delta$"): simulation after learning physical parameters (\ref{equ:theta}) except  geometry residual $\delta$. It shows incorrect contact locations due to inaccurate visual geometry, which causes object to "float" without contact. Third row ("with $\delta$"): simulation after learning all parameters including $\delta$. The result closely resembles real-world motion, capturing contact onset, rebound, and settling behavior, demonstrating Real2Sim's effectiveness in learning collision-accurate geometry and contact dynamics.

Despite visually appealing sim-real alignment in Fig. \ref{fig:dyn_adapt_tissue_box}, accurate long-horizon contact-rich scene prediction is still challenging due to chaotic/sensitive nature of contact physics. Minor variations in surface geometry can lead to large motion discrepancies. Also, small initial condition perturbations yield dramatically different outcomes. Thus, while we claim  Real2Sim enables the learning of an effective simulation model from visual inputs, it is best suited for one-step or few-step prediction. For longer horizons, it is preferable to integrate with real-time visual feedback to continually correct the accumulation of prediction errors, as in our Sim2Real robust tracking below.

\vspace{-1pt}
\subsection{Evaluation of Sim2Real}
\vspace{-3pt}
\begin{table*}[!t]
\centering
\renewcommand{\arraystretch}{1.2} 
\caption{Evaluation for Sim2Real robust tracking [cm].}
\resizebox{\linewidth}{!}{
\begin{tabular}{c|c|cccccccc|cc}
\hline
\textbf{Metrics} & \textbf{Method} & \textbf{snack box} & \textbf{tissue box} & \textbf{bottle} & \textbf{spam} & \textbf{duck} & \textbf{mustard} & \textbf{Leap Spam} & \textbf{Leap Duck} & \textbf{Avg.$\pm$Std.} \\ 
\hline
\multirow{5}{*}{\centering\rotatebox[origin=c]{90}{\small ADD$\downarrow$}}
& Baseline &2.20&2.14&3.58&3.54&2.66&7.13&1.88&1.01&3.01$\pm$1.86 \\
& w/o dynamics &1.54&9.50&20.29&5.83&3.54&4.81&6.75&5.45&7.21$\pm$5.76 \\
& random &>100.0&>100.0&>100.0&>100.0&1.83&2.74&1.70&3.22&>50.0 \\
& w/o $\delta$ &1.48&1.05&0.82&0.70&0.89&2.73&1.72&3.11&1.56$\pm$0.91 \\
& \bf{Proposed} &\bf{1.37}&\bf{1.04}&\bf{0.78}&\bf{0.67}&\bf{0.83}&\bf{2.62}&\bf{1.67}&\bf{0.87}& \bf{1.23$\pm$0.65} \\
\hline
\multirow{5}{*}{\centering\rotatebox[origin=c]{90}{\small ADD-S$\downarrow$}}
& Baseline &1.12&1.11&1.45&1.73&1.36&1.26&0.81&0.41&1.15$\pm$0.40 \\
& w/o dynamics &0.78&2.50&12.24&1.26&1.28&2.06&2.12&1.61&2.98$\pm$3.78 \\
& random &>100.0&>100.0&>100.0&>100.0&0.89&1.32&0.77&1.19&>50.0 \\
& w/o $\delta$ &0.79&0.55&0.40&0.34&0.49&1.31&0.77&1.31&0.74$\pm$0.38 \\
& \bf{Proposed} &\bf{0.74}&\bf{0.54}&\bf{0.39}&\bf{0.33}&\bf{0.39}&\bf{1.25}&\bf{0.75}&\bf{0.39}&\bf{0.59$\pm$0.31} \\
\hline
\end{tabular}
}
\vspace{-15pt}
\label{tab:tracking_evaluation}
\end{table*}

We evaluate Sim2Real for robust tracking in both scenarios and use {Average Distance of Model Points (ADD) and its symmetric variant (ADD-S)} metrics [cm] for tracking accuracy (Table~\ref{tab:tracking_evaluation}). As mentioned before, the pseudo  Ground-truth poses are obtained from FoundationPose \cite{wen2024foundationpose} with high-quality textured mesh models obtained from Scaniverse software. For comparison, we include a baseline, which is model-free tracking also using FoundationPose method~\cite{wen2024foundationpose} but fed with the same set of keyframes used in our Real2Sim.

We perform three ablation studies to assess the design contributions: "w/o dynamics" replaces contact dynamics with constant-velocity prediction (vision-only baseline); "random" uses contact dynamics with random physical parameters; "w/o $\delta$" learns all physical parameters except $\delta$ (visual geometry is used in contact dynamics model); "proposed" learns all parameters including $\delta$.

Table~\ref{tab:tracking_evaluation} shows the proposed method achieves best overall performance across all scenarios and objects. Learning contact physics and  geometry residual $\delta$ significantly outperforms all ablated variants. Baseline and "w/o dynamics" rely purely on visual tracking without contact physics, limiting performance under motion occlusions and blur. "w/o dynamics" introduces constant-velocity prior but lacks contact interaction prediction capability, causing drift/tracking errors during environment interaction. "w/o $\delta$" and "proposed" variants highlight the importance of accurate collision geometry for physics-aware tracking. The learned collision geometry residual $\delta$ effectively corrects visual geometry errors, leading to more accurate contact dynamics and higher Sim2Real tracking performance.

The tracking time cost for each module is summarized as follows. Experiments are conducted on a PC with a Ryzen 5955WX CPU and an RTX 4090 GPU. For each frame, SAM2 takes around 12ms, and SuperPoint  8ms. The  dynamics one-step prediction including neural SDF collision  completes in 5ms. Keyframe selection and LightGlue matching consume approximately 15ms, while the adaptive fusion optimization step adds an additional 5ms. Other  operations take about 3ms. In total, the system can achieve a processing rate of over 20Hz, enabling real-time performance for  6-DoF tracking. 

\vspace{-1pt}
\subsection{Failure Modes and Limitations}
\vspace{-5pt}
Our system can fail when visual observations are degraded or unavailable for extended periods. In particular, under prolonged vision missing, pose estimation becomes primarily driven by dynamics prediction; despite real2sim alignment, modeling errors can accumulate over long horizons and cause drift. The method also needs a reasonably accurate initial state to start: poor vision-based initialization may lead to incorrect collision geometry, which in turn degrades subsequent dynamics learning and prediction.

\section{CONCLUSIONS AND FUTURE WORK}
\vspace{-5pt}
We present \ourmethod{}, a physics-aware tracking framework bridging visual observations and contact physics for robust tracking of unknown dynamic objects in contact-rich environments. \ourmethod{} includes Real2Sim (builds contact dynamics model from visual observations through geometry estimation/refinement and physical property identification) and Sim2Real (uses learned contact dynamics to enhance 6-DoF pose visual tracking robustness/accuracy). Evaluation demonstrates robust real-time performance against occlusion and motion blur in highly dynamic contact-rich environments. 

In Real2Sim, we currently use a uniform residual, which cannot handle spatially varying errors in object geometry. We also assume a single rigid object and monocular RGB-D with a first-frame prompt, it cannot track multiple objects or deformable ones. In the future work, we will consider extending to multi-objects tracking scenarios and learning a spatial compensation that varies across the object's surfaces to enhance reconstruction accuracy. Also, our aligned Real2Sim environment may provide a more suitable training substrate for policies learning, potentially improving robustness and real-world transfer; we plan to investigate it in future work.

\addtolength{\textheight}{-1cm}   




\bibliographystyle{IEEEtran}  
\bibliography{IEEEfull}      

\end{document}